\pdfoutput=1
\documentclass[11pt]{article}

\usepackage{acl}

\usepackage{times}
\usepackage{latexsym}

\usepackage[T1]{fontenc}
\usepackage[utf8]{inputenc}

\usepackage{microtype}

\usepackage{inconsolata}

\usepackage[acronym]{glossaries}
\newacronym{rnn}{RNN}{Recurrent Neural Network}
\newacronym{emr}{EMR}{Electronic Medical Records}
\usepackage{comment}
\usepackage{subcaption}
\usepackage{graphicx}

\title{Visual Analytics for Generative Transformer Models % used in NLP %A Visual Analytics Framework for Supporting the Analysis of Generative Transformer Models for NLP
}

\author{Raymond Li\footnotemark[2]\footnotemark[1],~Ruixin Yang\footnotemark[2]\thanks{Co-first author},~Wen Xiao\footnotemark[2],\\
{\bf Ahmed AbuRa’ed\footnotemark[2],~ Gabriel Murray\footnotemark[3],~Giuseppe Carenini\footnotemark[2]}\\
\footnotemark[2]\hspace{.1em} University of British Columbia, Vancouver, BC, Canada \\
\footnotemark[3]\hspace{.1em} University of Fraser Valley, Abbotsford, BC, Canada \\
\texttt{\{\href{mailto:raymondl@cs.ubc.ca}{\texttt{raymondl}}, \href{mailto:ryang07@student.ubc.ca}{\texttt{ryang07}}, \href{mailto:xiaowen3@cs.ubc.ca}{\texttt{xiaowen3}}, \href{mailto:ahmed.aburaed@ubc.ca}{\texttt{ahmed.aburaed}},
\href{mailto:carenini@cs.ubc.ca}{\texttt{carenini}}\}@cs.ubc.ca} \\
\href{mailto:gabriel.murray@ufv.ca}{\texttt{gabriel.murray@ufv.ca}}
}

\begin{document}
\maketitle
\begin{abstract}
While transformer-based models have achieved state-of-the-art results in a variety of classification and generation tasks, their black-box nature makes them challenging for interpretability. In this work, we present a novel visual analytical framework to support the analysis of transformer-based generative networks. In contrast to previous work, which has mainly focused on encoder-based models, our framework is one of the first dedicated to supporting the analysis of transformer-based encoder-decoder models and decoder-only models for generative and classification tasks. Hence, we offer an intuitive overview that allows the user to explore different facets of the model through interactive visualization. To demonstrate the feasibility and usefulness of our framework, we present three detailed case studies based on real-world NLP research problems.
% makes analyzing them a challenging task.
% , namely abstractive summarization, machine translation, and question answering.
% common-sense QA.
\end{abstract}

\begin{figure*}[ht!]
  \includegraphics[width=\textwidth]{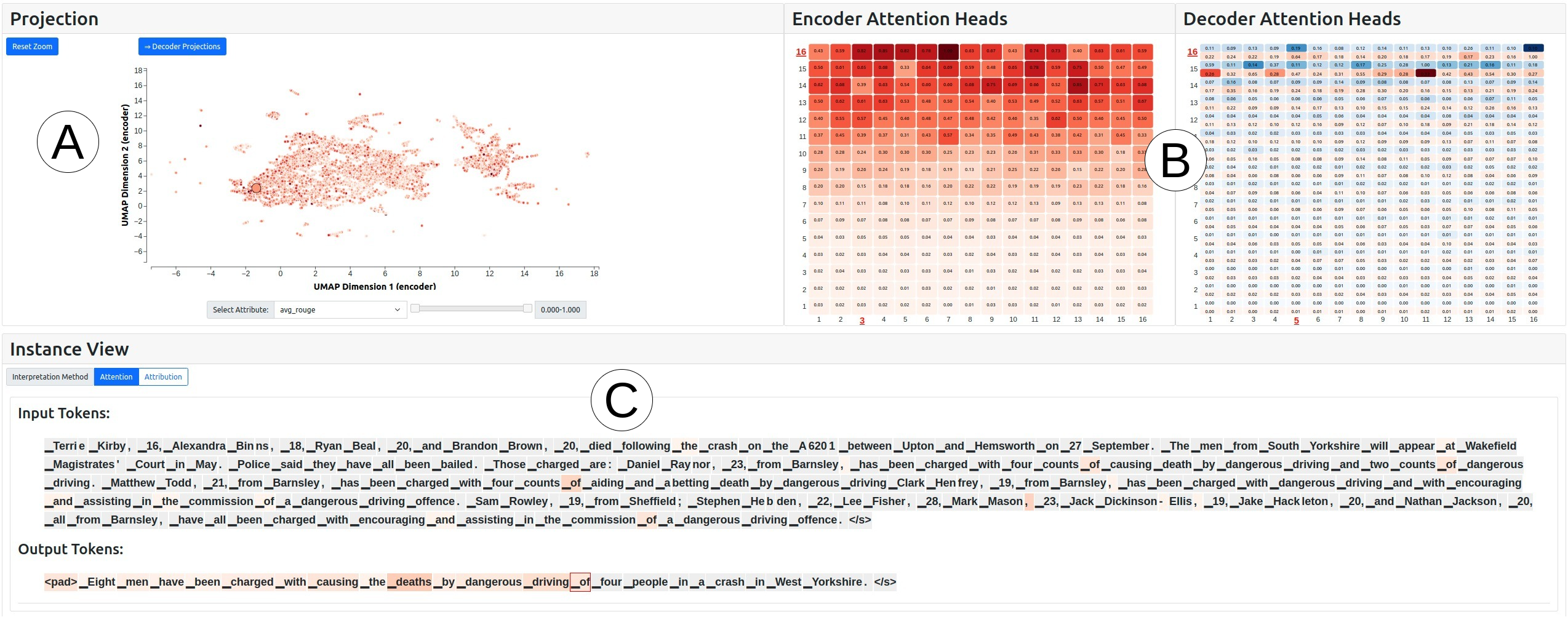}
  \caption{Overview of our interface: (A) Projection View uses a scatter-plot to visualize the hidden state projections of the corpus; (B) Attention Views visualizes the task importance of the encoder and decoder attention heads; (C) Instance View allows the user to analyze the input attribution and attention weights over the selected example.}
  % \Description{test}
  \label{fig:teaser}
  \vspace{-1em}
\end{figure*}

\section{Introduction}
Generative transformer models have demonstrated significant advancements in various tasks, including generation tasks such as summarization \cite{pegasus,xiao-etal-2022-primera}, machine translation \cite{liu-etal-2020-multilingual-denoising}, as well as classification tasks such as question answering and sentiment analysis \cite{min-etal-2022-rethinking}. There are two main kinds of architectures for the generative models, i.e. the encoder-decoder~\cite{lewis-etal-2020-bart,t5} and decoder-only models~\cite{gpt2,gpt3}. In the encoder-decoder model, the encoder computes the contextualized representation of the input sequence, which is subsequently used by a decoder to generate the output sequence in an auto-regressive manner. Conversely, in decoder-only models, input sequences are directly encoded as prompts,
% in the decoder, 
with text generation occurring by continuation from the given prompts.

While generative transformers have achieved state-of-the-art performance across a variety of benchmarks, there remain major drawbacks such as interpretability, 
% robustness,
performance on unseen examples (i.e. robustness),
and the reliability of quantitative metrics that prevent such models from being fully trusted. 
Moreover, there are challenges for specific NLP tasks that prevent such models from being safely deployed in the wild (e.g. hallucination for abstractive summarization), and the black-box nature of neural networks does not provide much insight for researchers to improve upon.
Therefore, two key questions that arise for a given model are: (i) what 
is the relative importance and contribution of each component to the final output sequence? (ii) does the model focus on specific contexts of the input sequence when making predictions? 
% How does the quality of the output sequence compare to the quantitative metric?
% How does the source representation change during inference? 
% What is the model's behaviour, facing an outlier as the input?

% the need for interactive visualization and brief intro on related works

Interactive visualizations %frameworks 
have been proposed
% systems have been proposed in the past 
to help researchers better understand the inner workings of neural models regarding the above questions. In essence, they %normally effectively 
encode abstract data and complex structures (e.g., embeddings and attention matrices) in meaningful graphical representations \cite{HohmanVAinDL} to facilitate human observations and qualitative evaluation.
% through interactions with controllable visual components, which provides feedback for model training and possible further improvements \cite{yuan2022isea}.
% To successfully answer those questions, an interactive visualization system would be helpful, as it can effectively encode abstract data and complex structures in meaningful graphical representations \cite{HohmanVAinDL} and facilitate human observations and evaluation through interactions with controllable visual components, which provides feedback for model training and possible further improvements. 
% Specifically, on transformers
% Specific
Specifically for transformers, there has been substantial work 
%Existing visual analytical systems for transformer models mainly focused 
on visualizing the hidden states dynamics and attention patterns \cite{vig2019multiscale, hoover-etal-2020-exbert}, and investigating model predictions with  instance-level saliency maps 
% attribution 
% and adversarial examples 
\cite{wallace2019allennlp, li2021t3}.
Tellingly, recent studies have shown that using the insights derived from interactive visualization can improve the model's performance by leveraging fixed attention maps discovered through the analysis process~\citep{li-etal-2022-human}.
% , 
% sevastjanova2021explaining},
% and on integrating both aspects in an end-to-end visualization framework \cite{tenney-etal-2020-language, li2021t3}. %that could provide insights for improving the performance of encoder-only transformers, such as BERT. 
However, most work has mainly focused on 
transformer encoders,
% possibly due to the earlier success of pretraining encoder-only models (e.g. \citep{devlin-etal-2019-bert, liu2019roberta, clark2020electra},
% encoder-only transformers,
with limited 
% attention paid to %there is no previous work
attention dedicated to
supporting the analysis of encoder-decoder and decoder-only architectures.
In this paper, we address this limitation %Our main research goal is to study how 
by proposing an interactive visual interface %that can be effectively %utilized% 
%used 
to support the development and interpretation of generative models, focusing in particular on NLP tasks such as abstractive summarization, machine translation, and question-answering.  
% To this end, we design the visual encodings and interactions based on the design goals (\autoref{sec:goals}) derived from existing literature.
% To this end, we carried out a design process (see \autoref{sec:goals}) loosely following the nested model of visualization \cite{5290695}, outlining detailed design goals and principles from a set of applicable user requirements, and designed the visual encodings and interactions accordingly. 
In summary, we make two contributions: % are:
% are two-fold:
%\begin{itemize}
%\item 
(i) we present the design and implementation of a visual analytical framework for transformer-based generative models (including encoder-decoder and decoder-only models), which incorporates hidden states dynamics, attention visualization, component importance, as well as input and output sequence attributions\footnote{\text{Code Repo: }\url{https://t.ly/ArcsU}}.
% Our framework comprises both the actual Visual Interface (in \autoref{sec:interface}) and the underlying algorithms (in \autoref{sec:algorithm}).
% To the best of our knowledge, our framework is the first to support the visualization of both the encoder-decoder and the decoder-only models (e.g. GPT series).
%\item 
(ii) we apply our framework to three important use cases with generative models (\autoref{sec:case-studies}), namely abstractive summarization, machine translation, and question answering, which shows that our system can be effectively used to support researchers in addressing open-research problems\footnote{Demo Video: \url{https://youtu.be/QmmdTIJpy1Y}}.
% the analysis of 
%\end{itemize}

% related work new draft version (based on Ahmed's version, focusing on more recent work)

\section{Related Work}
\label{sec:related}
% Given the current limited understanding of how transformer-based language models function as black-box models despite their significant progress in numerous tasks, various interactive visualization systems have been proposed to aid in the investigation of model explanations. 

Due to the popularity of the pretrain-finetune paradigm \citep{devlin-etal-2019-bert}, most prior works have focused on the transformer encoder architecture. For examples, 
BertViz~\citep{vig-2019-multiscale} visualized the self-attention weights in transformer encoder models while exBERT~\citep{hoover2019exbert} extended their work to support the analysis of hidden state representations. LMExplorer \citep{sevastjanova2021explaining} adopted self-similarity score \citep{ethayarajh-2019-contextual} to visualize the degree of contextualization for the hidden states, and DODRIO \citep{wang2021dodrio} added the comparison of attention weights to syntactic structure and semantic information. Other frameworks such as AllenNLP \citep{wallace2019allennlp} and Language Interpretability Tool \citep{tenney-etal-2020-language} have integrated input attribution to visualize the local explanation for the model's prediction, while T3-Vis \citep{li2021t3} also incorporated the training dynamics to support the fine-tuning process.

Meanwhile, %few research efforts have 
less research has been devoted to supporting the analysis of generative transformer models, %Most visual frameworks have focused on support 
with most proposals focusing on the visualization of the predicted output. For example, LMdiff~\citep{strobelt2021lmdiff} visualized the qualitative differences between language models based on output distributions, while LM-Debugger~\citep{geva-etal-2022-lm} supported the interactive modification of a
%used an interactive approach to modify 
feed-forward network %(FFN) 
for controllable text generation. The %most related work
work %that is 
most closely related to ours is Ecco \citep{alammar-2021-ecco}, %which is 
a library for visualizing 
model components,
% and analyzing various components of BERT-based and GPT2-based models, 
including input saliency, 
hidden states, and neuron activations.
% However, their system focuses on the ranking and similarity of the hidden states in %while 
examining the attribution for a single example, while %our framework 
we allow %the flexibility for he 
users to interactively select and analyze different examples in the corpus.
% However, instead of a visual framework, they provide a Python package that runs inside Jupyter notebooks, with function components for instance-level inspections. In contrast, our end-to-end system supports the loading and examination of an entire dataset, and has an interactive interface that allows users without deeper knowledge of the model to navigate different components easily.
However, unlike the visual framework we propose, Ecco is a Python package designed for Jupyter notebooks, offering the capability for instance-level inspections, whereas our interactive interface enables the dynamic exploration of model behaviour on a complete dataset and allows for the easy navigation of users without deep knowledge of the model. 

\begin{figure*}[ht!]
    \centering
    \includegraphics[width=.7\textwidth]{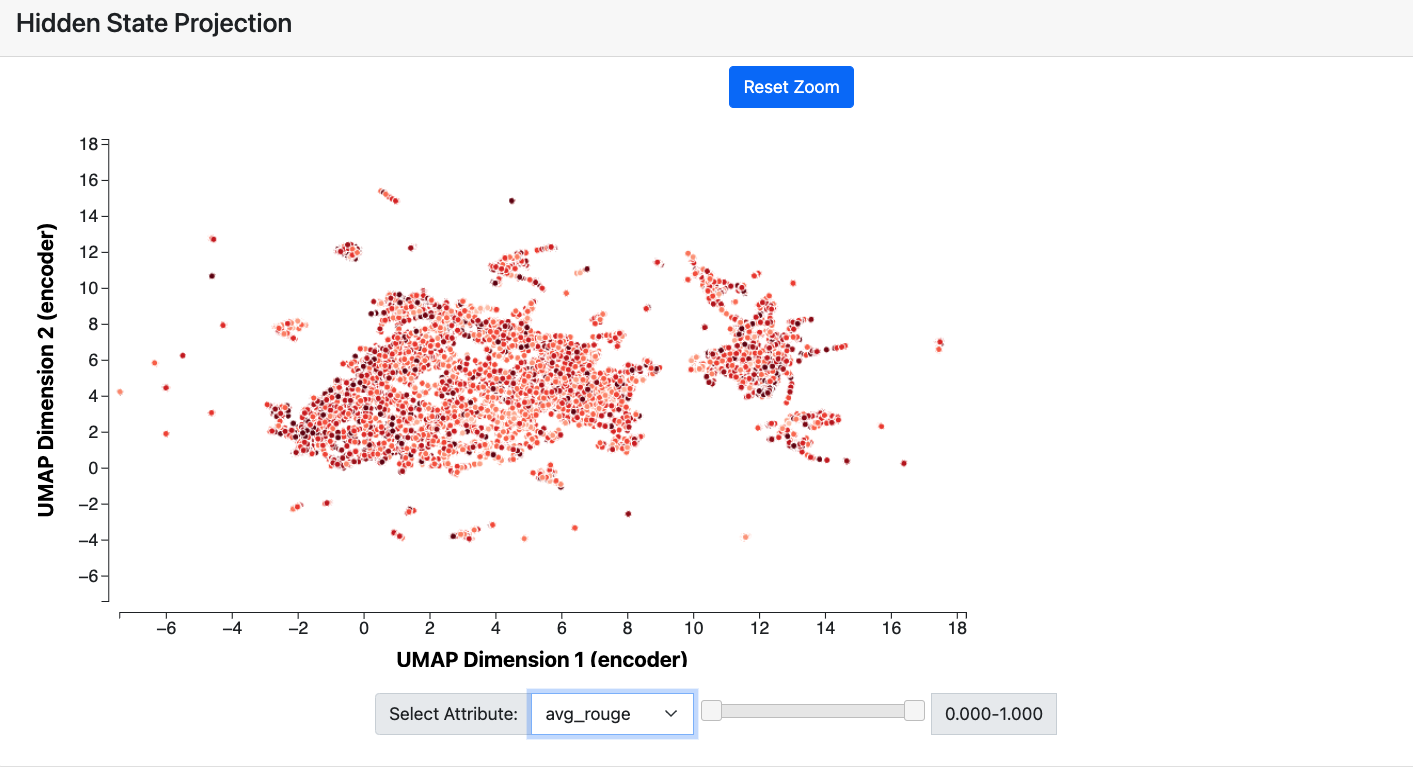}
    \caption{In the \textbf{Projection View}, each point on the scatter-plot encodes a document in the XSum corpus for the task of abstraction summarization.}
    \label{fig:projection-view}
    \vspace{-.7em}
\end{figure*}

\section{Visual Interface}
\label{sec:interface}
As shown in \autoref{fig:teaser}, our interface is divided into three views to support model exploration at different levels of granularity. In this section, we describe the visual components in detail.
% and justify the design and interactions based on the design goals described in \autoref{sec:goals}.

\subsection{Projection View}
\label{sec:projection-view}
The Projection View (\autoref{fig:projection-view}) provides an overview of the hidden state representations, where each point on the scatter-plot encodes a single example in the corpus, projected to a two-dimensional space using 
% UMAP 
the UMAP~\citep{mcinnes2018umap} or t-SNE~\citep{van2008visualizing} algorithm (Details in Appendix \ref{sec:umap}) with user adjustable parameters. % and T-SNE%. 
In order to visualize the representations for the decoding steps, we employ an ``overview+detail'' technique \citep{cockburn2009review}. When the user clicks on a point on the scatter-plot, the detailed view of the corresponding example visualizes the projection of decoder hidden states for all time steps of the output token sequence. This view also serves as the primary method for selecting examples to visualize.
% (G1).

%The user can also encode any continuous attribute (e.g. length, evaluation metric) with color

We provide the option to use color for encoding any continuous attribute (e.g. length, evaluation metric) selected by the users. In the example displayed in \autoref{fig:projection-view},  color encodes the average ROUGE-score \citep{lin-2004-rouge} of the document compared with the ground-truth summary. We also provide a range-selector for the user to filter the scatter-plot based on the selected attribute to better support the selection of examples for analysis.

% \subsection{Data Table}
% In the Data Table View, we implement a scrollable list that allows the user to search for examples based on keywords (G1). We further extend the search function with word replacement, allowing the user to test for model robustness on perturbed examples (G3).

\subsection{Attention Views}
\label{sec:attention-view}

\begin{figure*}[ht!]
% \centering
% \begin{subfigure}
%   \centering
  \includegraphics[width=.5\linewidth]{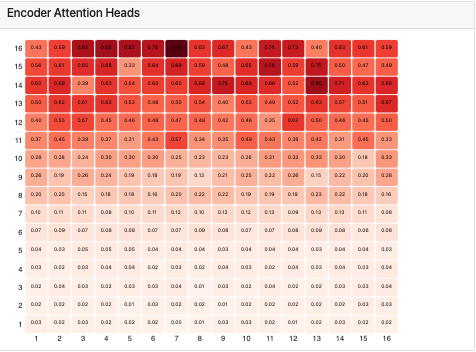}
  % \caption{A subfigure}
  % \label{fig:encoder-heads}
% \end{subfigure}%
% \begin{subfigure}
%   \centering
  \includegraphics[width=.5\linewidth]{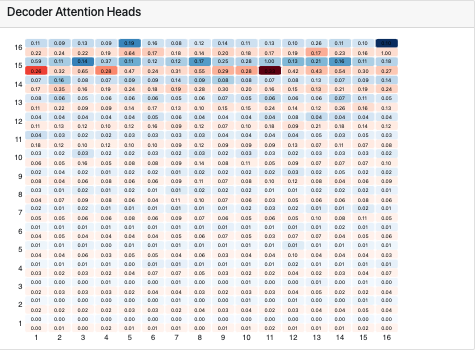}
  % \caption{A subfigure}
%   \label{fig:sub2}
% \end{subfigure}
\caption{The \textbf{Attention Views} for the encoder (left) and decoder (right) heads, the color saturation of each heatmap cell encodes the task important of the corresponding attention head.}
\label{fig:attention-view}
\end{figure*}

In the Attention Views (\autoref{fig:attention-view}), we provide an overview visualization of the model's attention head importance (of the encoder attentions for the encoder-decoder models and decoder attentions for both architectures).
% (G2)
% . 
Specifically, for both transformer encoder and decoder with $l$ hidden layers and $h$ attention heads, we encode the head importance score \citep{hao2021self} (Details in Appendix \ref{sec:head-importance}) using a $l \times h$ heatmap where the color saturation of each cell encodes the importance of the corresponding attention head w.r.t the prediction. Each cell of the encoder heatmap represents one attention head, while it is divided into two subcells in the decoder heatmap, with the top subcell (blue) encoding the importance of the cross-attention head and the bottom subcell (red) encoding the decoder attention head.
By selecting a heatmap cell, the user can visualize the corresponding attention distribution 
% on the selected example 
in the Instance View.%(\autoref{sec:instance-view}). 
 
\subsection{Instance View}
\label{sec:instance-view}

\begin{figure*}[ht]
    \centering
    \includegraphics[width=.8\linewidth]{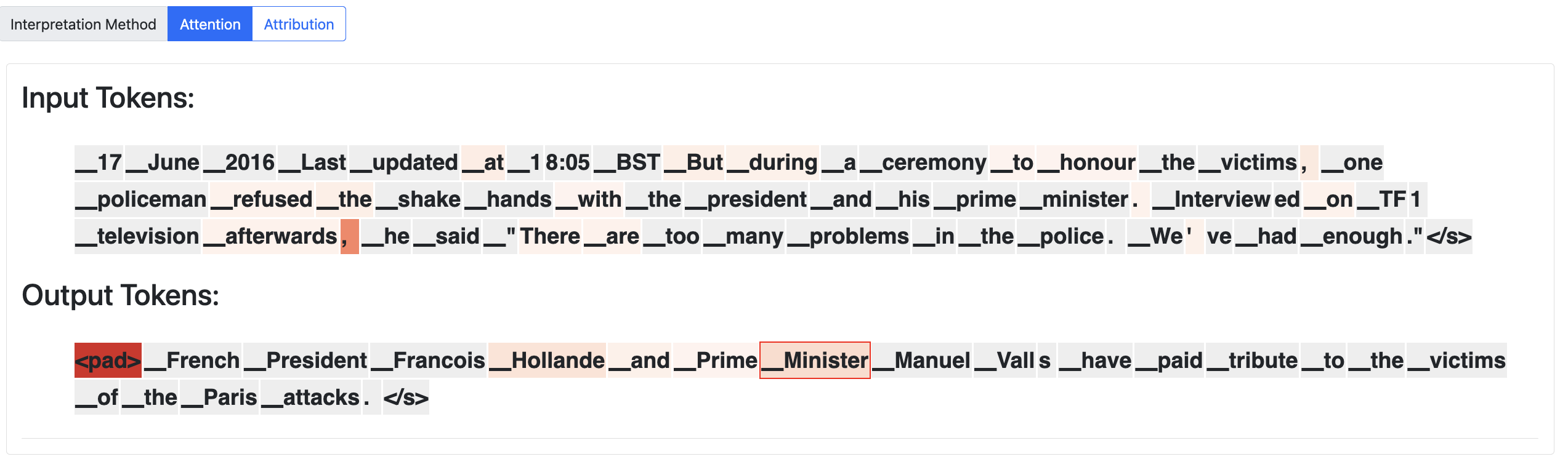}
    \caption{In the \textbf{Instance View}, the input/output tokens are separately encoded as one-dimensional heatmaps. The above example visualizes the decoder self and cross-attention distribution of the output token ``Minister''. }
    \label{fig:instance-view}
    \vspace{-1em}
\end{figure*}

Lastly, the Instance View (\autoref{fig:instance-view}) of our interface assists the instance-level investigation of the selected data example 
% (G1)
. We use two heatmaps to separately visualize the document input and model output. In particular, each token is encoded as a cell in the heatmap, where we provide two interpretation methods to analyze the model's behaviour on the selected example. First, we provide a visualization of the attention distribution over the selected example (\autoref{sec:projection-view}) and attention heads (\autoref{sec:attention-view}), allowing the user to examine the specific behaviour of the attention weights. 
% (G2). 

To scale for the length of the input and output, we use a one-dimensional heatmap over the input/output tokens 
that wraps around the width of the container,
and allow the user to select the attention matrix row by clicking on individual tokens. 
For encoder-decoder models, clicking on an input token selects the corresponding row in the encoder self-attention matrix, while clicking on an output token selects the corresponding row in the decoder self-attention and cross-attention matrix. For autoregressive decoder-only models, we use only the input token container, where clicking on the token selects the row in the decoder self-attention matrix.

Further, we provide an alternative method to visualize model predictions 
% (G3) 
using the gradient-based input attribution \citep{sundararajan2017axiomatic} and interaction scores \citep{janizek2021explaining}. The attribution and interaction score estimates the relative importance and pairwise interactions of the input tokens w.r.t. the model predictions (Details in Appendix \ref{sec:input-attribution} and \ref{sec:input-interactions}).
% described in \autoref{sec:input-attribution} and \autoref{sec:input-interactions}. 
When the user selects a token, the heatmap will be used to encode either the token-level attribution score of the input (and prior output tokens for encoder-decoder) at the current generation step, or the pairwise interaction score with all other tokens summed over all generation steps.
\vspace{.5em}
% Similar to the importance in Component View, the token-level importance is also estimated using gradient-based attribution methods. Users are able to compare token attributions of different methods from a side-by-side view. %Finally, we use a stacked bar chart to display the relative contributions of the source and target tokens at the final time step, based on LRP \cite{voita-etal-2021-analyzing}.%

\begin{figure}[h!]
    \centering
    \vspace{-1em}
    \includegraphics[width=.5\textwidth]{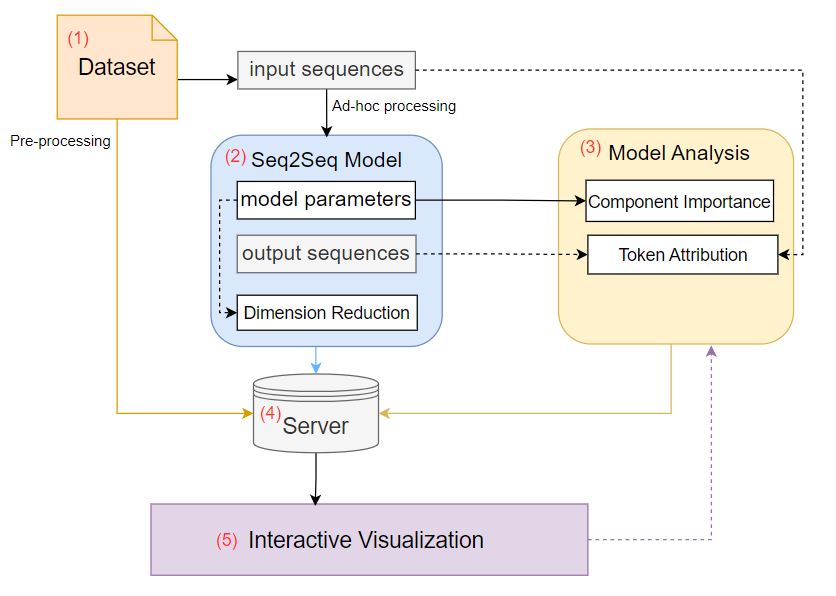}
    \caption{Overview of our system.}
    \vspace{-2em}
    \label{fig:system-architecture}
\end{figure}
% \vspace{-1em}

\begin{figure*}[ht!]
% \centering
% \begin{subfigure}
%   \centering
  \includegraphics[width=.5\linewidth]{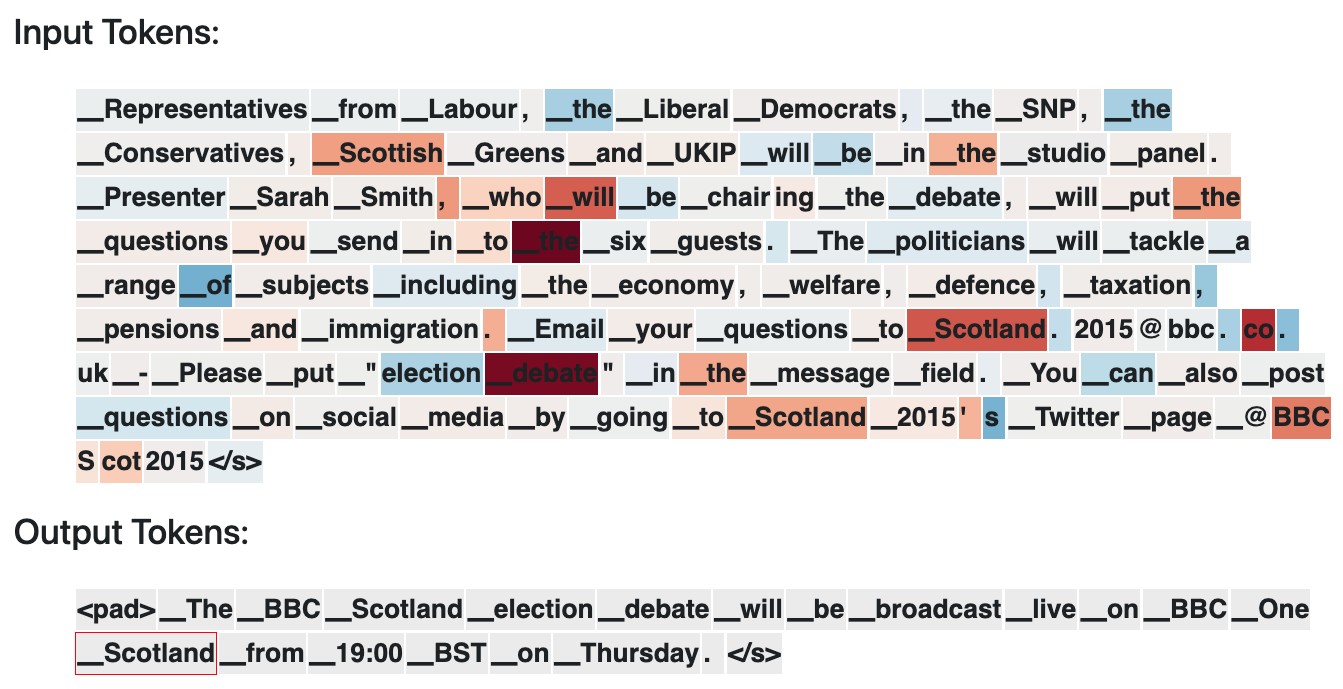}
  % \caption{A subfigure}
  % \label{fig:encoder-heads}
% \end{subfigure}%
% \begin{subfigure}
%   \centering
  \includegraphics[width=.5\linewidth]{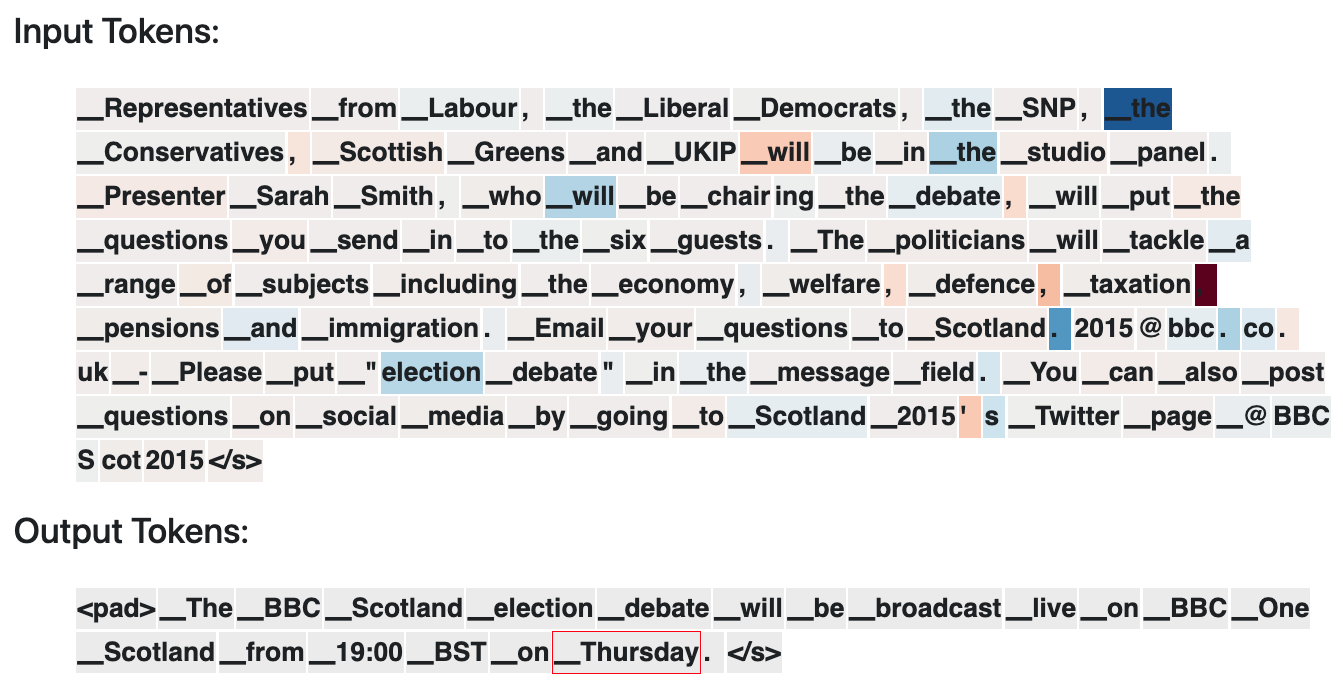}
  % \caption{A subfigure}
%   \label{fig:sub2}
% \end{subfigure}
\caption{An example of the input attribution for a truthful entity (left) and a hallucinated entity (right), with blue representing the negative attributes and red representing the positive attributes. It can be found that there are more negative input attributions in the hallucinated entity. }
\label{fig:case-study-1}
\vspace{-.6em}
\end{figure*}

% \section{System Architecture and Implementation}
\section{System Overview}
\label{sec:system}
\autoref{fig:system-architecture} displays the data processing workflow and general architecture of our system, which comprises five modules (numbered accordingly as in the figure): (1) the dataset loader module that loads the required datasets from the Datasets library \citep{lhoest-etal-2021-datasets}; (2) the model loader module that loads the pre-trained models from the Transformers library \citep{wolf-etal-2020-transformers} and performs dimensionality reduction on hidden state projections; (3) the model analysis module that implements different component importance and token attribution methods, extending components from the Captum package \citep{kokhlikyan2020captum}; (4) the Flask-based backend server that processes relevant resources (model parameters, projections, and output sequences) that are needed for the interface; (5) the frontend interface (see \autoref{sec:interface}) implemented using JavaScript and D3.js \citep{bostock2011d3} for interactive visualization.

\section{Case Studies}
\label{sec:case-studies}
In order to provide evidence for the usability and usefulness 
% value
of our interface,
% as well as to evaluate our design goals proposed in \autoref{sec:goals}, 
we present three case studies that attempt to address open-research problems for generative models in key NLP tasks. 

\subsection{Entity-level Hallucination in Abstractive Summarization}

% \begin{figure}[ht!]
%     \centering
%     \includegraphics[width=.6\textwidth]{figures/system-architecture.JPG}
%     \caption{Overview of our system.}
%     \label{fig:system-architecture}
% \end{figure}

% In the first study, we analyze the hallucination behaviour for abstractive summarization models, where the generated summaries contain unfaithful content to the source documents.
Abstractive Summarization generates summaries for a given document using transformer-based encoder-decoder architecture. However, generated summaries often suffer from factual inconsistencies known as "hallucinations", which can occur on either the entity or relation level~\cite{nan-etal-2021-entity}. Recent works have explored ways to detect and reduce these inconsistencies \cite{cao-etal-2022-hallucinated}
% ,spancopy}.

In this case study, we explore the entity-level factual inconsistencies in PEGASUS model-generated summaries on the Xsum dataset \cite{narayan-etal-2018-dont}. Previous studies have shown that this SOTA model produces over $30\%$ hallucinated entities \cite{pegasus}. The goal is to gain insight into the types and properties of hallucinated entities generated by the model and use this understanding to 
% enhance
improve summarization performance.

The user first applies an entity-level factual inconsistency score~\cite{nan-etal-2021-entity} as the criteria to select source documents with summaries containing more hallucinated entities in the projection view. In the Instance View, it is observed that the generated summaries often contain hallucinations such as the full name of a person instead of just the first or last name. The input token attributions of these hallucinated entities are compared with those of truthful entities in the Instance View, and it is found that the tokens with higher contribution to the hallucinated entities tend to be stop words and have more negative input attributions. Additionally, the model appears to be less focused when generating hallucinated tokens as indicated by the higher entropy over the input attributions. These findings are consistent with previous works \cite{cao-etal-2022-hallucinated}, which showed that hallucinated entities are generated with lower probabilities.

Overall, in this case study the user has successfully explored the hallucinated entities by using our interface, and the findings suggest further research ideas on improving the factualness of the summarization model. For instance, a threshold could be applied on the entropy of the input attributions to filter out the potentially nonfactual tokens.

\subsection{Attention Patterns in Neural Machine Translation}
The current state-of-the-art models for the machine translation task are generally based on the encoder-decoder transformer architecture \citep{liu-etal-2020-multilingual-denoising}, where text in the source language (e.g., English) is mapped into text in the target language (e.g., Chinese). %Despite  however, the explanations of the model's behavior are still under explored.
% The multi-head attention mechanism is an integral component of the transformer model, which is currently the state-of-the-art architecture for neural machine translation. 
While abundant previous work on supporting the interpretability of such machine translation models has investigated attention weights for positional or syntactic patterns \citep{vig-belinkov-2019-analyzing, clark-etal-2019-bert, kovaleva-etal-2019-revealing, huber-carenini-2022-towards}, often with the aid of visualization tools, most of these studies were focused only on the self-attention matrices of the transformer encoders. In this second case study, 
inspired by \citet{voita-etal-2019-analyzing} who identified attention head roles with ground-truth annotations, the user employs our visual interface to examine interpretable patterns in the encoder and cross-attention of machine translation models. The goal is to gain insights into the information processed by the encoder and the alignment between the source and target text.
% motivated by the work of \citet{voita-etal-2019-analyzing} that characterized the roles of attention heads with ground-truth annotations, the user applies our visual interface looking for interpretable patterns in both the encoder attention and the cross attention of the machine translation models, to better understand what information is considered in the encoder and the alignment between the source and target text. 
In this study, we use the OPUS-MT \citep{tiedemann-thottingal-2020-opus} model (6-layer, 8-head) trained on an English-to-Chinese corpus \footnote{https://github.com/Helsinki-NLP/Tatoeba-Challenge/blob/master/models/eng-zho/README.md}.

To analyze the transformer-based language model, the user first filters documents by input length (Projection View) and selects the important attention heads (Attention Head View), where the user discovers that only a few attention heads are significant in both the encoder and decoder. By using the Instance View, the important heads focus on either the local information (e.g. the previous or subsequent token, or the tokens within the same sentence boundary) or the matching information in the context, i.e. the tokens with similar meaning to the current token. The cross-attention heads provide alignment information between input and target tokens, either on the token level (attending to source tokens with exactly the same meaning) or on the sentence level (attending to tokens in the corresponding sentence of the input sequence). The findings are consistent with prior works and could lead to further research on explainable translation models. The small number of important attention heads suggests potential for parameter pruning ~\cite{michel2019sixteen}, or by directly injecting patterns into the model~\citep{li-etal-2022-human}.

\subsection{In-Context Learning for Multi-Choice Question Answering}

Recent studies on large language models (mostly decoder-only models) have shown their ability to perform a task by simply prepending a few input-label pairs as demonstrations of the input examples~\cite{gpt3}, which is also referred to as in-context learning. However, despite their impressive performance on a wide array of downstream tasks \citep{liu2023pre}, the exact behaviour of these models remains unclear. 
% While the performance of in-context learning generally improves as the model scales up, performing analysis with smaller models may allow us to gain a better understanding of the model's behavior due to the feasibility of computation (e.g. computing gradients), and the gained insights can be applied to larger model variants.

In this case study, we perform an analysis of GPT-2 Large \citep{gpt2} on the multi-choice CommonsenseQA dataset \citep{talmor-etal-2019-commonsenseqa}.
% \footnote{https://github.com/Alrope123/rethinking-demonstrations}. 
Specifically, we adopt the MetaICL variant \citep{min-etal-2022-metaicl}, where the GPT-2 is first fine-tuned to perform in-context learning on a large set of training tasks. During inference, the model computes the perplexity for each question-answer pair and predicts the pair with the lowest perplexity as the answer. We focus on the input interactions (\autoref{sec:input-interactions}) between input tokens to understand why the model assigns a low perplexity for incorrect examples. 

\begin{figure}[ht!]
     \centering
     \begin{subfigure}[b]{\linewidth}
         \centering
         \includegraphics[width=\textwidth]{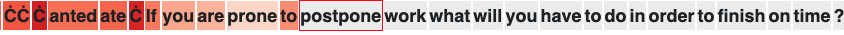}
     \end{subfigure}
     \hfill
     \begin{subfigure}[b]{\linewidth}
         \centering
         \includegraphics[width=\textwidth]{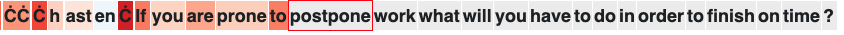}
     \end{subfigure}
    \caption{Interaction score between false positive prediction ``antedate'' and false negative prediction ``antedate''.}
    \label{fig:case-study-3}
\end{figure}

Initially, in the Attention Head View, the user notices that a few heads in the lower layers have higher importance scores, with three of the top five most important heads in the first layer. Using the Projection View, the user filters by false positive examples and then selects examples with the highest relative loss. Since the in-context demonstration examples are fixed during inference, the interaction (see \autoref{sec:input-interactions}) between question and answer tokens is analyzed using the Instance View. The user finds that the model displays a similar pattern-matching behaviour as in previous case studies, where semantically related tokens have high interaction scores. For instance, in \autoref{fig:case-study-3}, the token "postpone" has a high interaction score with the false-positive answer "antedate" and a low interaction score with the ground-truth answer "hasten". We hypothesize that while the demonstrations help `locate' a previously learned concept to do the in-context learning task, answering questions still relies on the correlations with the pretraining data.
\section{Conclusions and Future Work}
In this work, we propose a visual analytics framework for transformer-based generative models that allows the exploration of the model and corpus across multiple levels of granularity while providing the user with faceted visualization of the model's behaviour 
through dynamic interactions. To demonstrate the usability of our framework, we present three case studies based on open research problems in NLP. We hope the findings will motivate future studies on using our framework to address challenges in other tasks and larger pre-trained models.
While our framework is under active development, we will keep refining it by exploring additional usage scenarios and incorporating user feedback.

% As the framework is under active development,  we will continue to improve
% our framework through the iterative process of exploring additional usage scenarios and collecting feedback from users.
% (e.g. question answering). 

\bibliography{anthology, custom}
% \bibliographystyle{acl_natbib}

% \newpage
\appendix

\section{Algorithm}
\label{sec:algorithm}
In this section, we describe the algorithms utilized in our interface for data transformation and model analysis. 

\subsection{Dimensionality Reduction}
\label{sec:umap}
The goal of dimension reduction is to reduce high-dimensional data (e.g. hidden states) into low-dimensional space while retaining some meaningful properties (e.g. distance) of the original data. In our interface, we adopt UMAP~\citep{mcinnes2018umap}, which is a non-linear reduction method that preserves the global structure. 
We also provide the user with an alternative t-SNE~\citep{van2008visualizing} projection for contrast. 
Specifically, we apply the dimension reduction method to visualize the averaged encoder hidden states for encoder-decoder models, and the average decoder hidden states for decoder-only models.

\subsection{Attention Head Importance}
\label{sec:head-importance}
To estimate the task importance of individual attention heads, we adapt the Attention Attribution method proposed by \citep{hao2021self} for the three types of attention heads, namely, encoder self-attention, decoder self-attention, and decoder cross-attention.
Specifically, for the weights of the $j$th attention head $A_j$, its attribution score is computed as:
\begin{equation}
    \textrm{Attr}(A_j) = A_j \odot \int_{\alpha=0}^{1}\frac{\partial L(\alpha A)}{\partial A_j}d\alpha
\label{eq:head-attr}
\end{equation}
where $\odot$ is the element-wise multiplication operator, and $L$ can either be the loss function for the task (e.g. cross-entropy) or the loss w.r.t to the predicted logit. Intuitively, $\textrm{Attr}(A_j)$ considers how sensitive model predictions are to the attention weights. In practice, the line integral can be efficiently computed using the Riemann approximation where the gradients are summed at points occurring at small intervals along the path from the zero matrices (no information) to the attention weights ($A_i$). The importance score is averaged across all examples in the corpus. 

\subsection{Input Attribution}
\label{sec:input-attribution}
In order to compute the attribution of input token $x_i$ at each prediction step, we use the Integrated Gradients (IG) method \citep{sundararajan2017axiomatic} where:
\begin{equation}
    Attr(x_i) =  x_i \odot \int_{\alpha=0}^1 \frac{\partial F(\alpha x)}{\partial x_i}d\alpha
\end{equation}
where $F$ is the transformer model. Note that this is almost identical to the Attention Attribution in \autoref{eq:head-attr}. However, since we are interested in understanding the attribution to the model prediction, the gradient is computed w.r.t. the predicted logit rather than the loss function.

\subsection{Input Interactions}
\label{sec:input-interactions}
Input interactions refer to the degree of interactions between input tokens in affecting the model's predictions. While many existing techniques have used second-order derivatives to estimate the pairwise interaction score \citep{janizek2021explaining}, we find this to be infeasible for larger models due to the memory requirements for computing the Hessian. Rather, we use the approach for estimating the head importance \citet{hao2021self}, where we sum up the attention attribution score over the attention heads to obtain the pairwise interaction scores between tokens.

\end{document}